\documentclass[10pt,twocolumn,letterpaper]{article}

\usepackage{wacv}
\usepackage{times}
\usepackage{epsfig}
\usepackage{graphicx}
\usepackage{amsmath}
\usepackage{amssymb}
\usepackage{subcaption}
\usepackage{multirow}

\usepackage[pagebackref=true,breaklinks=true,letterpaper=true,colorlinks,bookmarks=false]{hyperref}
\usepackage{cleveref}

\wacvfinalcopy 

\def\wacvPaperID{811} 

\begin{document}

\title{Action Segmentation with Mixed Temporal Domain Adaptation}

\author{Min-Hung Chen$^1$\thanks{Work done during an internship at Baidu USA} \hspace{1em}
Baopu Li$^2$ \quad
Yingze Bao$^2$ \quad
Ghassan AlRegib$^1$\\
$^1$Georgia Institute of Technology \quad
$^2$Baidu USA\\
}

\maketitle

\begin{abstract}
   The main progress for action segmentation comes from densely-annotated data for fully-supervised learning. Since manual annotation for frame-level actions is time-consuming and challenging, we propose to exploit auxiliary unlabeled videos, which are much easier to obtain, by shaping this problem as a domain adaptation (DA) problem. 
   Although various DA techniques have been proposed in recent years, most of them have been developed only for the spatial direction. Therefore, we propose \textbf{Mixed Temporal Domain Adaptation (MTDA)} to jointly align frame- and video-level embedded feature spaces across domains, and further integrate with the domain attention mechanism to focus on aligning the frame-level features with higher domain discrepancy, leading to more effective domain adaptation. Finally, we evaluate our proposed methods on three challenging datasets (GTEA, 50Salads, and Breakfast), and validate that MTDA outperforms the current state-of-the-art methods on all three datasets by large margins (e.g. 6.4\% gain on F1$@50$ and 6.8\% gain on the edit score for GTEA).
\end{abstract}

\section{Introduction}
\begin{figure}[!t]
\centering
\includegraphics[width=0.475\textwidth]{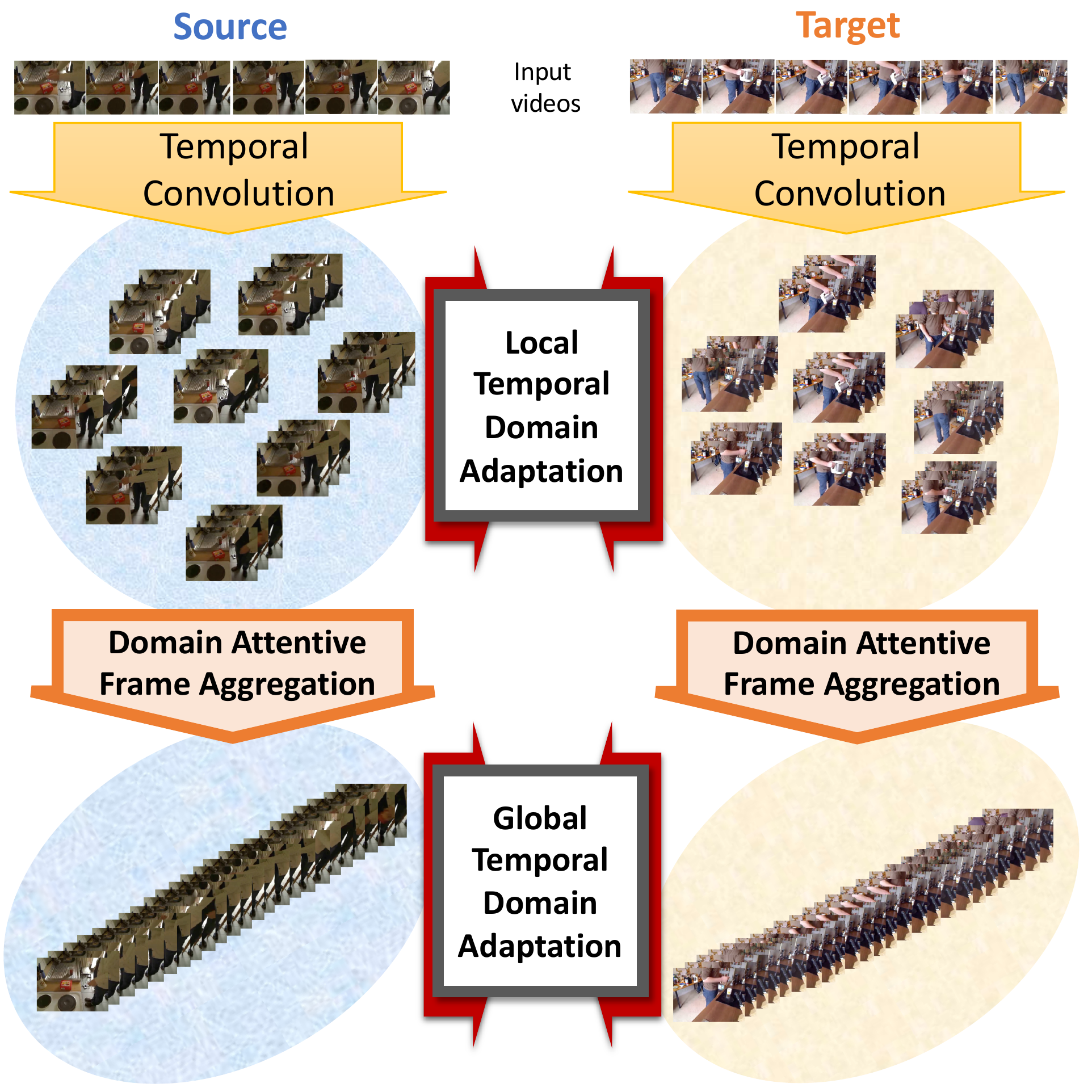}
\caption{An overview of the proposed MTDA for action segmentation. ``Source" refers to the data with labels, and ``Target" refers to the unlabeled data with the standard transductive setting for DA. We first extract local temporal features using temporal convolution, and then obtain global temporal features with domain attentive frame aggregation. Finally, we diminish the domain discrepancy by jointly performing local and global temporal domain adaptation. 
Here we use the video \textit{making tea} as an example.}
\label{fig:overview_action_seg_DA}
\end{figure}

Action segmentation is of significant importance for a wide range of applications, including video surveillance and analysis of daily human activities. Given a video, the goal is to simultaneously segment the video by time and predict each segment with a  corresponding action category.
While video classification has shown great progress given the recent success of deep neural networks~\cite{wang2018non, ma2018attend, ma2019ts}, temporally locating and recognizing action segments in long untrimmed videos is still challenging.

Action segmentation approaches can be factorized into extracting low-level features using convolutional neural networks and applying high-level temporal models. 
Encouraged by the advances in speech synthesis~\cite{oord2016wavenet}, recent approaches rely on temporal convolutions to capture long
range dependencies across frames using a hierarchy of temporal convolutional filters~\cite{lea2017temporal, ding2017tricornet, lei2018temporal, farha2019ms}. 

Despite the success of these temporal models, the performance gains come from densely-annotated data for fully-supervised learning. Since manually annotating precise frame-by-frame actions is time-consuming and challenging, these methods are not easy to extend to larger scale for real-world applications.

In this paper, we regard action segmentation as a domain adaptation (DA) problem with the transductive setting~\cite{pan2010survey, csurka2017comprehensive} given the observation that the main challenge is the distributional discrepancy caused by spatio-temporal variations across domains. For example, different people (also noted as \textit{subjects}) may perform the same action with different styles in terms of spatial locations and temporal duration. The variations in the background environment also contribute to the overall domain discrepancy. To solve this problem, we propose to diminish the domain discrepancy by utilizing auxiliary unlabeled videos, which are much easier to obtain. 

Videos can suffer from domain discrepancy along both the spatial and temporal directions, bringing the need of alignment for embedded feature spaces along both directions~\cite{chen2019temporal}. However, most DA approaches have been developed only for images and not videos~\cite{long2015learning, long2017deep, ganin2015unsupervised, ganin2016domain, li2017revisiting, li2018adaptive, saito2018maximum}. Therefore, we propose \textbf{Mixed Temporal Domain Adaptation (MTDA)} to jointly align frame- and video-level embedded feature spaces across domains, as shown in \Cref{fig:overview_action_seg_DA}. We further integrate with the domain attention mechanism to focus on aligning the frame-level features with higher domain discrepancy, leading to more effective domain adaptation. 
To support our claims, we evaluate our approaches on three datasets with high spatio-temporal domain discrepancy: Georgia Tech Egocentric Activities (GTEA)~\cite{fathi2011learning}, 50Salads~\cite{stein2013combining}, and the Breakfast dataset~\cite{kuehne2014language}, and achieve new state-of-the-art performance on all three datasets. 
Since our approach can adapt a model trained in one environment to new environments using only unlabeled videos without additional manual annotation, it is applicable to large-scale real-world scenarios, such as video surveillance.

In summary, our contributions are three-fold:
\begin{enumerate}
    \item \textbf{Local Temporal Domain Adaptation}: 
    We propose an effective adversarial-based DA method to learn domain-invariant frame-level features. To the best of our knowledge, this is the first work to utilize unlabeled videos as auxiliary data to diminish spatio-temporal variations for action segmentation. 
    
    \item \textbf{Mixed Temporal Domain Adaptation (MTDA)}:
    We jointly align local and global embedded feature spaces across domains by integrating additional DA mechanism which aligns the video-level feature spaces. Furthermore, we integrate the domain attention mechanism to aggregate domain-specific frames to form global video representations, leading to more effective domain adaptation. 
    
    \item \textbf{Experiments and Analyses}: 
    We evaluate on three challenging real-world datasets and outperform all the previous state-of-the-art methods. We also perform analysis and ablation study on different design choices to identify key contributions of each component.

\end{enumerate}

\section{Related Works} 
In this section, we review the most recent work for action segmentation including the fully- and weakly-supervised setting. We also review the most related domain adaptation work for images and videos.

\noindent\textbf{Action Segmentation.}
Encouraged by the advances in speech synthesis~\cite{oord2016wavenet}, recent approaches rely on temporal convolutions to capture long-range dependencies across frames using a hierarchy of temporal convolutional filters~\cite{lea2017temporal, ding2017tricornet, lei2018temporal, farha2019ms}. 
ED-TCN~\cite{lea2017temporal} follows an encoder-decoder architecture with a temporal convolution and pooling in the encoder, and upsampling followed by deconvolution in the decoder. 
TricorNet~\cite{ding2017tricornet} replaces the convolutional decoder in the ED-TCN with a bi-directional LSTM (Bi-LSTM). 
TDRN~\cite{lei2018temporal} builds on top of ED-TCN~\cite{lea2017temporal} and use deformable convolutions instead of the normal convolution and add a residual stream to the encoder-decoder model. 
MS-TCN~\cite{farha2019ms} stacks multiple stages of temporal convolutional network (TCN) where each TCN consists of multiple temporal convolutional layers performing acausal dilated 1D convolution. 
With the multi-stage architecture, each stage takes an initial prediction from the previous stage and refines it. We build our approach on top of MS-TCN, focusing on developing methods to effectively exploit unlabeled videos instead of modifying the architecture.

\noindent\textbf{Domain Adaptation.}
Most recent DA approaches are based on deep learning architectures designed for addressing the domain shift problems given the fact that the deep CNN features without any DA method outperform traditional DA methods using hand-crafted features~\cite{donahue2014decaf}. Most DA methods follow the two-branch (source and target) architecture, and aim to find a common feature space between the source and target domains. The models are therefore optimized with a combination of classification and domain losses~\cite{csurka2017comprehensive}. 

One of the main classes of methods used is \textit{Discrepancy-based DA}, whose metrics are designed to measure the distance between source and target feature distributions, including variations of maximum mean discrepancy (MMD)~\cite{long2015learning, long2016unsupervised, zellinger2017central, yan2017mind, long2017deep} and the CORAL function~\cite{sun2016deep}. By diminishing the distance of distributions, discrepancy-based DA methods reduce the gap across domains. 
Another common method, \textit{Adversarial-based DA}, adopts a similar concept as GANs~\cite{goodfellow2014generative} by integrating domain discriminators into the architectures. Through the adversarial objectives, the discriminators are optimized to classify different domains, while the feature extractors are optimized in the opposite direction. ADDA~\cite{tzeng2017adversarial} uses an inverted label GAN loss to split the optimization into two parts: one for the discriminator and the other for the generator. In contrast, the gradient reversal layer (GRL) is adopted in some works~\cite{ganin2015unsupervised, ganin2016domain, zhang2018collaborative} to invert the gradients so that the discriminator and generator are optimized simultaneously. 

Recently, TADA~\cite{wang2019transferable} adopts the attention mechanism to adapt the transferable regions and images. Differently, 
we design the attention mechanism for spatio-temporal domains, aiming to attend to the important parts of temporal dynamics for domain adaptation.

\noindent\textbf{Domain Adaptation for Actions.}
Unlike image-based DA, video-based DA is still an under-explored area. A few works focus on small-scale video DA with only few overlapping categories~\cite{sultani2014human, xu2016dual, jamal2018deep}. \cite{sultani2014human} improves the domain generalizability by decreasing the effect of the background. \cite{xu2016dual} maps source and target features to a common feature space using shallow neural networks. AMLS~\cite{jamal2018deep} adapts pre-extracted C3D~\cite{tran2015learning} features on a Grassmann manifold obtained using PCA. However, the datasets used in the above works are too small to have enough domain shift to evaluate DA performance. 
Recently, Chen et al.~\cite{chen2019temporal} propose two larger cross-domain datasets for action recognition and the state-of-the-art approach TA$^3$N. 
However, these works focus only on the classification task while we concentrate on the more challenging temporal segmentation task.

\section{Technical Approach}
We first introduce our baseline model which is the current state-of-the-art approach for action segmentation, MS-TCN~\cite{farha2019ms} (\Cref{sec:baseline}). And then we describe how we incorporate unlabeled video to align frame-level feature spaces (\Cref{sec:local_DA}), and present our proposed method with the attention-based video-level domain adaptation (\Cref{sec:MTDA}).

\subsection{Baseline Model: MS-TCN} \label{sec:baseline}
\begin{figure}[!t]
\centering
\includegraphics[width=0.475\textwidth]{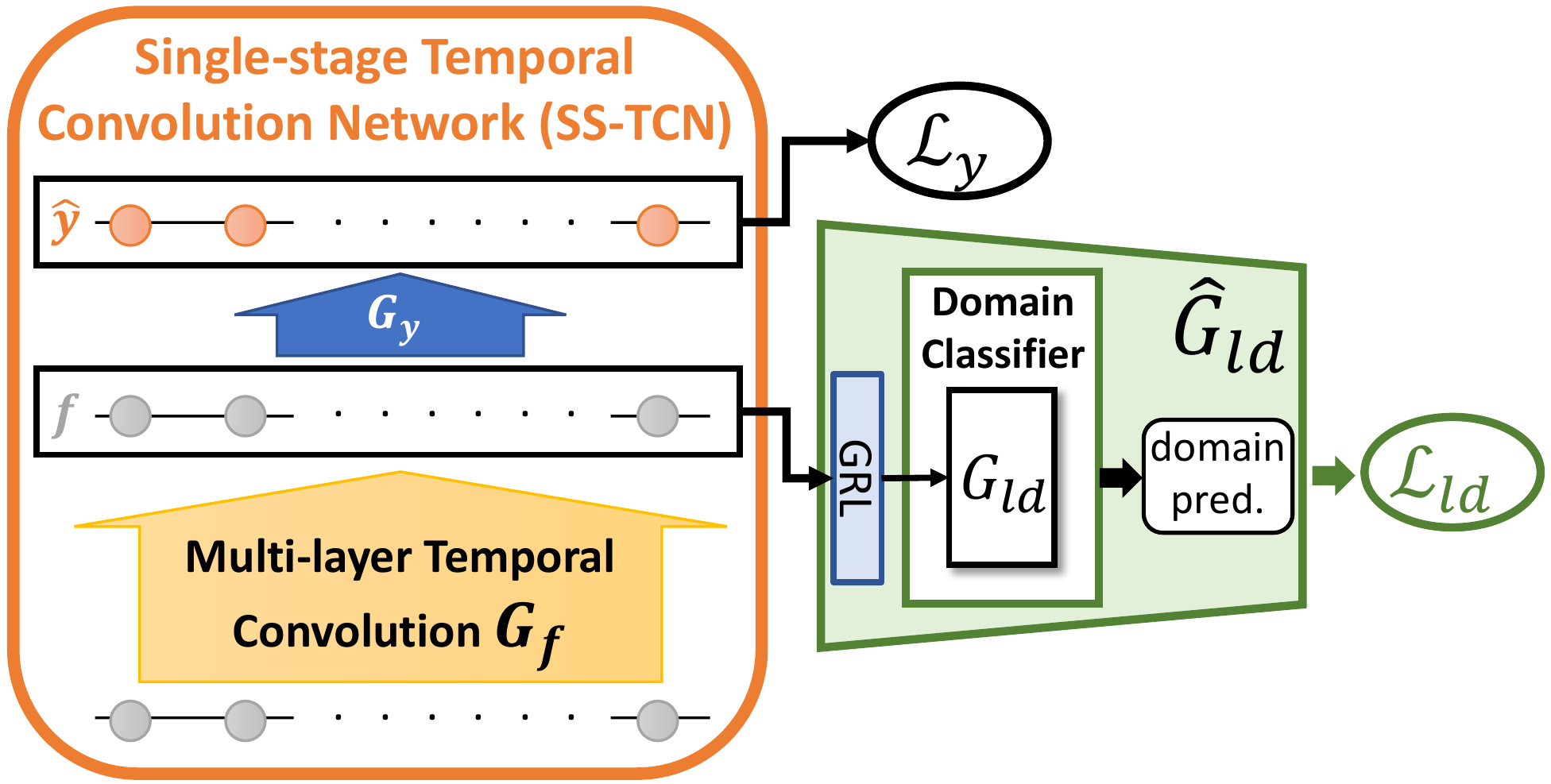}
\caption{We perform local temporal DA by applying the domain classifier $G_{ld}$ to the final embedded features $\textit{\textbf{f}}$ in one stage. A gradient reversal layer (GRL) is added between $G_{ld}$ and $\textit{\textbf{f}}$ so that $\textit{\textbf{f}}$ is trained to be domain-invariant.
$\boldsymbol{\hat{y}}$ is the frame-level predictions for each stage. $\mathcal{L}_y$ and $\mathcal{L}_{ld}$ are the prediction loss and local domain loss, respectively. 
}
\label{fig:local_temporal_DA}
\end{figure}

The basic component of our baseline model is single-stage temporal convolutional network (SS-TCN), as shown in the left part of \Cref{fig:local_temporal_DA}. SS-TCN consists of multiple temporal convolutional layers performing acausal dilated 1D convolution. Dilated convolution is used to increase the temporal receptive field exponentially without the need to increase the number of parameters, which can prevent the model from over-fitting the training data.    
Motivated by the success of multi-stage architectures~\cite{wei2016convolutional, newell2016stacked}, several SS-TCNs are stacked to form the multi-stage TCN (MS-TCN). Each stage takes the prediction from the previous stage and utilizes the multi-layer temporal convolution feature generator $G_f$ to obtain the frame-level features $\textit{\textbf{f}}=\{f_1, f_2, ..., f_T\}$, and then converts them into the frame-level predictions $\mathbf{\hat{y}}=\{{\hat{y}}_1, {\hat{y}}_2, ..., {\hat{y}}_T\}$ 
by a fully-connected layer $G_y$.   

The overall prediction loss function for each stage is a combination of a classification loss and a smoothing loss, which can be expressed as follows:
\begin{equation} \label{eq:loss_baseline}
\small
\mathcal{L}_y = \mathcal{L}_{cls} + \alpha\mathcal{L}_{T-MSE}
\end{equation}
where $\mathcal{L}_{cls}$ is a cross-entropy loss,  $\mathcal{L}_{T-MSE}$ is a truncated mean squared error used to reduce the difference between adjacent frame-level prediction to improve the smoothness, and $\alpha$ is the trade-off weight for the smoothness loss. To train the complete
model, we minimize the sum of the losses over all stages.

\subsection{Local Temporal Domain Adaptation} \label{sec:local_DA}
Despite the progress of MS-TCN on action segmentation, there is still a large room for improvement. The main challenge is the distributional discrepancy caused by spatio-temporal variations across domains. For example, different subjects may perform the same action completely different due to personalized spatio-temporal styles. Therefore, the problem becomes generalizing the model across domains. In this paper, we propose to reduce the domain discrepancy by performing unsupervised DA with auxiliary unlabeled videos.

Encouraged by the success of adversarial-based DA approaches~\cite{ganin2015unsupervised, ganin2016domain}, for each stage, we feed the frame-level features $\textit{\textbf{f}}$ to an additional shallow binary classifiers, called the \textit{local domain classifiers $G_{ld}$}, to discriminate whether the data is from the source or target domain.
Before back-propagating the gradients to the main model, a gradient reversal layer (GRL) is inserted between $G_{ld}$ and the main model to invert the gradient, as shown in Figure~\ref{fig:local_temporal_DA}. 
During adversarial training, $G_f$ is learned by maximizing the domain discrimination loss $\mathcal{L}_{ld}$, while $G_{ld}$ is learned by minimizing $\mathcal{L}_d$ with the domain label $d$. 

Therefore, the feature generator $G_f$ will be optimized to gradually align the feature distributions between the two domains. 
In this paper, we note the \textit{adversarial local domain classifier $\hat{G}_{ld}$} as the combination of a GRL and a domain classifier $G_{ld}$, and investigate the integration of $G_{ld}$ for different stages. From our experiments, the best performance happens when $G_{ld}$s are integrated into middle stages. For more details, please see \Cref{sec:stage_analysis}.

The overall loss function becomes a combination of the baseline prediction loss $\mathcal{L}_y$ and the local domain loss $\mathcal{L}_{ld}$, which can be expressed as follows:
\begin{equation} \label{eq:loss_baseline-localdomain}
\small
\begin{split}
\mathcal{L} = \sum^{N_s}\mathcal{L}_y - \sum^{\widetilde{N_s}}\beta_l\mathcal{L}_{ld}
\end{split}
\end{equation}
\begin{equation} \label{eq:loss_localdomain}
\small
\mathcal{L}_{ld} = \frac{1}{T}\sum_{j=1}^{T}L_{ld}(G_{ld}(f_j),d_j)
\end{equation}
where $N_s$ is the total stage number, $\widetilde{N_s}$ is the number of selected stages, and $T$ is the number of frames from each video. $\textit{L}_{ld}$ is a binary cross entropy loss function, and $\beta_l$ is the trade-off weight for local domain loss $\mathcal{L}_{ld}$.

\subsection{Mixed Temporal Domain Adaptation} \label{sec:MTDA}
One main drawback of integrating DA into local frame-level features $\textit{\textbf{f}}$ is that the video-level feature space is still not fully aligned. Although $\textit{\textbf{f}}$ is learned using the context and dependencies from neighbor frames, the temporal receptive field still not guarantees to cover the whole video length. 
Furthermore, aligning video-level feature spaces also helps to generate domain-adaptive frame-level predictions for action segmentation.
Therefore, we propose to jointly align local frame-level feature spaces and global video-level feature spaces, as shown in \Cref{fig:MTDA}.

\noindent\textbf{Global Temporal Domain Adaptation.}
To achieve this goal, we first aggregate $\textit{\textbf{f}}=\{f_1, f_2, ..., f_T\}$ using temporal pooling to form video-level feature $V$. Since each feature $f_t$ captures context in different time by temporal convolution, $V$ still contain temporal information despite the naive temporal pooling method. After obtaining $V$, we add another domain classifier (noted as \textit{global domain classifier $G_{gd}$}) to explicitly align the embedded feature spaces of video-level features.

Therefore, the global domain loss $\mathcal{L}_{gd}$ is added into the overall loss, which can be expressed as follows:
\begin{equation} \label{eq:loss_baseline-localdomain-globaldomain}
\small
\begin{split}
\mathcal{L} = \sum^{N_s}\mathcal{L}_y - \sum^{\widetilde{N_s}}(\beta_l\mathcal{L}_{ld}+\beta_g\mathcal{L}_{gd})
\end{split}
\end{equation}
\begin{equation} \label{eq:loss_globaldomain}
\small
\mathcal{L}_{gd} = L_{gd}(G_{gd}(G_{tf}(\boldsymbol{f})),d)
\end{equation}
where $\textit{L}_{gd}$ is also a binary cross entropy loss function, and $\beta_g$ is the trade-off weight for global domain loss $\mathcal{L}_{gd}$. 

\begin{figure}[!t]
\centering
\includegraphics[width=0.475\textwidth]{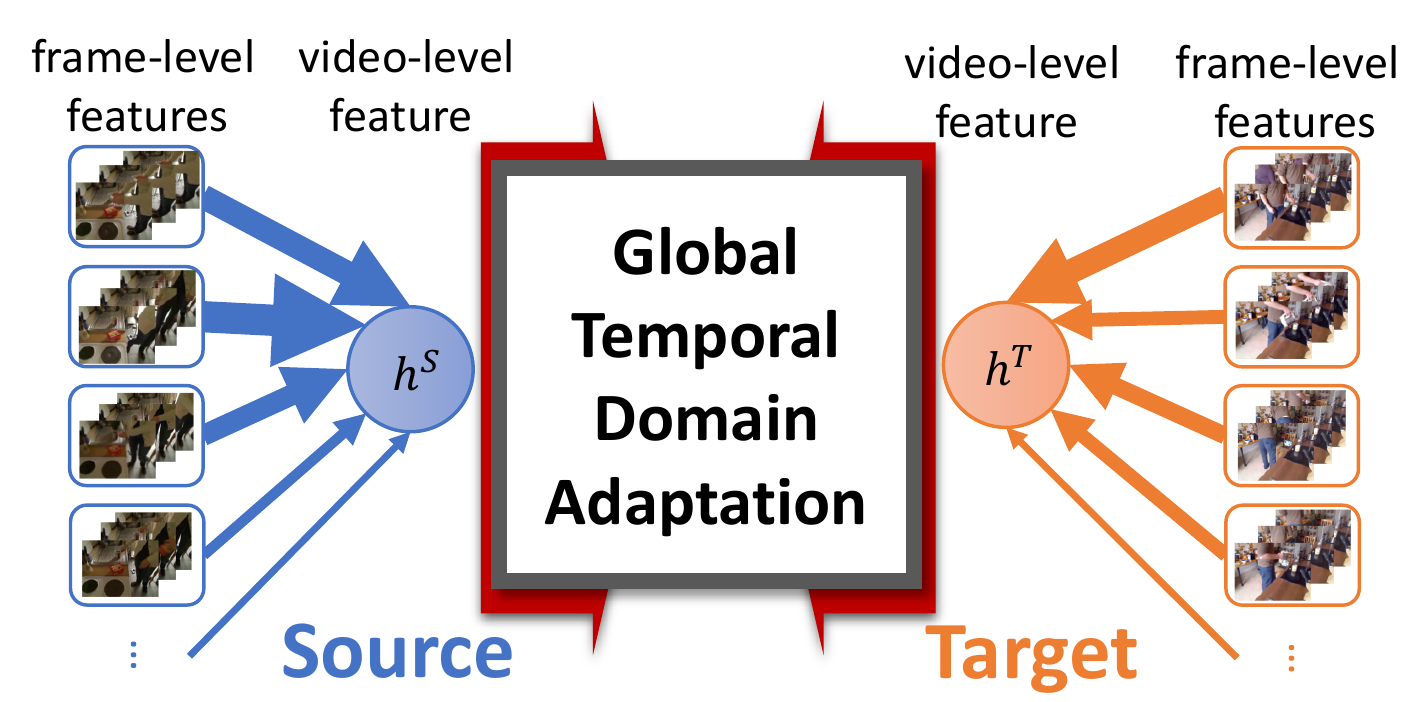}
\caption{The overview of global temporal DA with the domain attention mechanism. Frame-level features are aggregate with different attention weights to form the video-level feature $h$ for global domain DA.  
Thicker arrows corresponds to larger attention weights.
}
\label{fig:overview_domain_attention}
\end{figure}

\noindent\textbf{Domain Attention Mechanism.}
Although aligning video-level feature spaces across domains benefits action segmentation, not all the frame-level features are equally important to align. In order to effectively align overall temporal dynamics, we want to focus more on aligning the frame-level features which have larger domain discrepancy. Therefore, we assign larger attention weights to those features, as shown in \Cref{fig:overview_domain_attention}.

\begin{figure}[!t]
\centering
\includegraphics[width=0.475\textwidth]{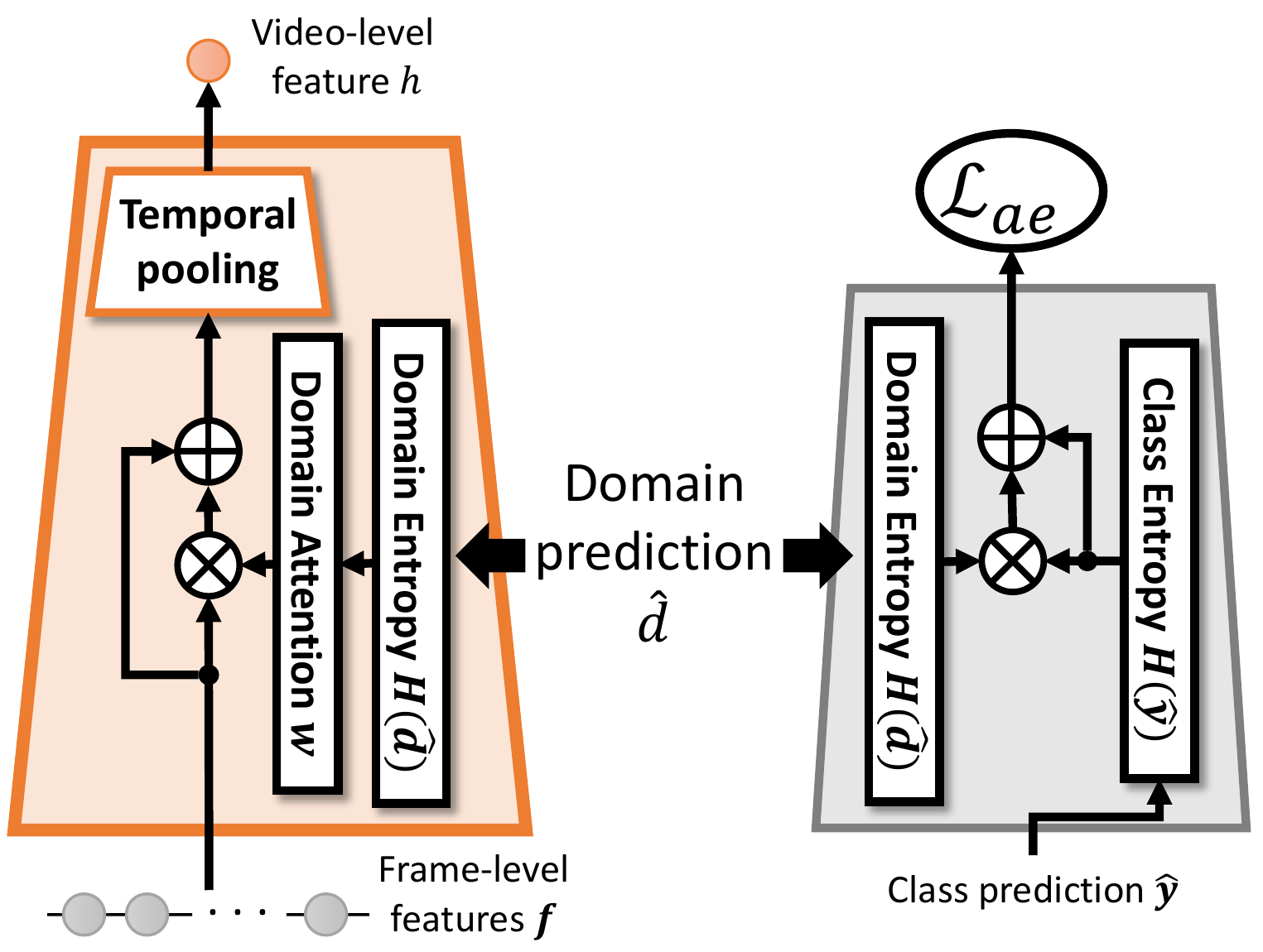}
\caption{The details of the domain attention mechanism, consisting of two modules: domain attentive temporal pooling (\textit{left}) and domain attentive entropy (\textit{right}). Both modules use the domain prediction $\hat{d}$ to make their inputs domain attentive with a residual connection. $\mathcal{L}_{ae}$ is the attentive entropy loss.}
\label{fig:domain_attn_full}
\end{figure}

\begin{figure*}[!t]
\centering
\includegraphics[width=\textwidth]{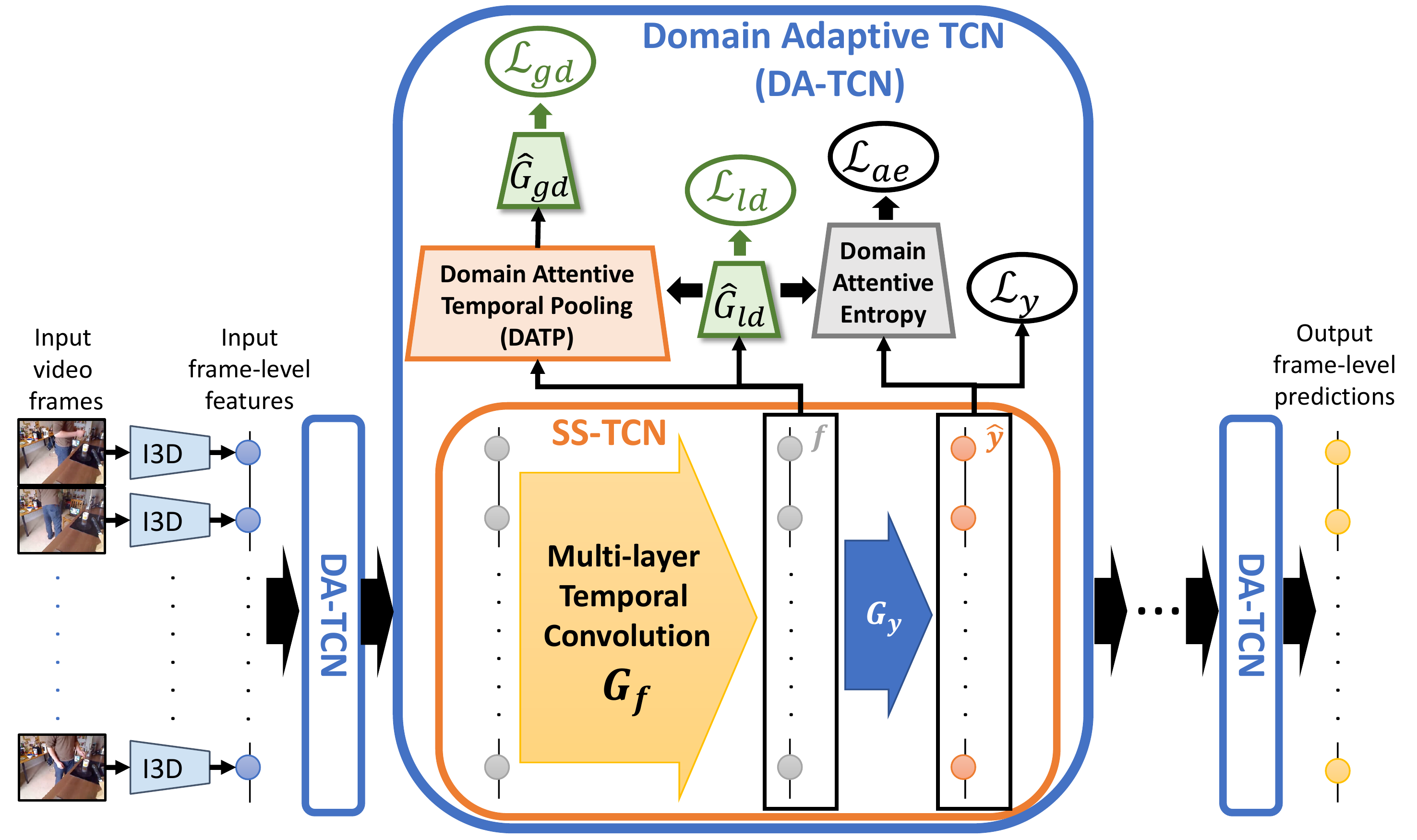}
\caption{The overall architecture of the proposed MTDA. By equipping with a local adversarial domain classifier $\hat{G}_{ld}$, a global adversarial domain classifier $\hat{G}_{gd}$, and the domain attention mechanism as shown in \Cref{fig:domain_attn_full}, we convert a SS-TCN into a domain adaptive TCN (DA-TCN), and stack multiple stages of DA-TCN to build the final architecture. $\mathcal{L}_{ld}$ and $\mathcal{L}_{gd}$ is the local and global domain loss, respectively. $\mathcal{L}_{y}$ is the prediction loss and $\mathcal{L}_{ae}$ is the attentive entropy loss.}
\label{fig:MTDA}
\end{figure*}

Hence, we integrate each stage with the domain attention mechanism, as shown in \Cref{fig:domain_attn_full}, which utilizes the entropy criterion to generate the domain attention value for each frame-level feature as below:
\begin{equation} \label{eq:attention-weight}
\small
w_j = 1 - H(\hat{d}_j)
\end{equation}
where $\hat{d}_j$ is the domain prediction from $G_{ld}$. $H(p) = -\sum_{k}p_k\cdot\log(p_k)$ is the entropy function to measure uncertainty. $w_j$ increases when $H(\hat{d}_j)$ decreases, which means the domains can be distinguished well.
We also add a residual connection for more stable optimization. Finally, we aggregate the attended frame-level features with temporal pooling to generate the video-level feature $h$, which is noted as \textit{domain attentive temporal pooling (DATP)} and can be expressed as:
\begin{equation} \label{eq:attention-residual}
\small
h = \frac{1}{T}\sum_{j=1}^{T}(w_j + 1) \cdot f_j
\end{equation}

In addition, we add the minimum entropy regularization to refine the classifier adaptation. However, we only want to minimize the entropy for the videos that are similar across domains. 
Therefore, we attend to the videos which have low domain discrepancy, so that we can focus more on minimizing the entropy for these videos. 
The attentive entropy loss $\mathcal{L}_{ae}$ can be expressed as follows:
\begin{equation} \label{eq:loss_attentive-entropy}
\small
\mathcal{L}_{ae}=\frac{1}{T}\sum_{j=1}^{T}(1+H(\hat{d}_j)) \cdot H(\hat{y}_j)
\end{equation}
where $\hat{d}$ and $\hat{y}$ is the output of $G_{ld}$ and $G_y$, respectively. We also adopt the residual connection for stability.

By adding \Cref{eq:loss_attentive-entropy} into \Cref{eq:loss_baseline-localdomain-globaldomain}, and replacing $G_{tf}(\boldsymbol{f})$ with $h$ by \Cref{eq:attention-residual}, 
the overall loss of our final proposed \textbf{Mixed Temporal Domain Adaptation (MTDA)}, as shown in \Cref{fig:MTDA}, can be expressed as follows:
\begin{equation} \label{eq:loss_MTDA}
\small
\begin{split}
\mathcal{L} = \sum^{N_s}\mathcal{L}_y - \sum^{\widetilde{N_s}}(\beta_l\mathcal{L}_{ld}+\beta_g\mathcal{L}_{gd}-\mu\mathcal{L}_{ae})
\end{split}
\end{equation}
where $\mu$ is the weight for the attentive entropy loss.

\section{Experiments}
To evaluate how our approaches can diminish spatial-temporal discrepancy for action segmentation, we choose three challenging datasets: Georgia Tech Egocentric Activities (GTEA)~\cite{fathi2011learning}, 50Salads~\cite{stein2013combining}, and the Breakfast dataset~\cite{kuehne2014language}, which separate the training and testing sets by different people (noted as \textit{ subjects}), resulting in high spatio-temporal variations. By following the transductive setting for DA, ``Source" refers to the original training set, and ``Target" refers to the testing set without labels. With these three datasets, we show how our approaches adapt the same actions across different people by decreasing the spatio-temporal variations across videos.

\subsection{Datasets}
The \textbf{GTEA} dataset contains 28 videos including 7 kitchen activities performed by 4 subjects. All the videos were recorded by a camera that is mounted on the actor’s head. 
There are totally 11 action classes including background. On average, each video is around one minute long with 20 action instances. 
We use 4-fold cross-validation for evaluation by leaving one subject out.

The \textbf{50Salads} dataset contains 50 videos for salad preparation activities performed by 25 subjects. There are totally 17 action classes. On average, each video contains 20 action instances and is 6.4 minutes long. 
For evaluation, we apply 5-fold cross-validation by leaving five subjects out.

The \textbf{Breakfast} dataset is the largest among the three datasets with 1712 videos for breakfast preparation activities performed by 52 subjects. The videos were recorded in 18 different kitchens with 48 action classes where each video contains 6 action instances on average and is around 2.7 minutes long. For evaluation, we utilize the standard 4-fold cross-validation by leaving 13 subjects out.

\subsection{Evaluation Metrics}
For all the three datasets, we use the following
evaluation metrics as in \cite{lea2017temporal}: frame-wise accuracy (Acc), segmental edit score, and segmental F1 score at the IoU threshold $k\%$, denoted as F1$@k$ ($k=\{10,25,50\}$). 

While \textbf{frame-wise accuracy} is one of the most common evaluation metrics for action segmentation, it does not take into account the temporal dependencies of the prediction, causing large qualitative differences with similar frame-wise accuracy. In addition, long action classes have higher impact on this metric than shorter action classes, making this metric not able to reflect over-segmentation errors. 
To address the above limitations, the \textbf{segmental edit score} penalizes over-segmentation by measuring the ordering of predicted action segments independent of slight temporal shifts.
Finally, another suitable metric \textbf{segmental F1 score (F1$\boldsymbol{@k}$)} becomes popular recently since it is found that the score numbers better indicate the qualitative segmentation results. F1$@k$ also penalizes over-segmentation errors while ignoring minor temporal shifts between the predictions and ground truth. F1$@k$ is determined by the total number of actions but not depends on the duration of each action instance, which is similar to mean average precision (mAP) with intersection-over-union (IoU) overlap criteria.

\subsection{Implementation}
Our implementation is based on the PyTorch~\cite{paszke2017automatic} framework. 
We extract I3D~\cite{carreira2017quo} features for the video frames and use these features as input to our model. The video frame rates are the same as \cite{farha2019ms}.
For fair comparison, we adopt the same architecture design choices of MS-TCN~\cite{farha2019ms} as our baseline model. 
The whole model consists of four stages where each stage contains ten dilated convolution layers. We set the number of filters to $64$ in all
the layers of the model and the filter size is $3$. 
For optimization, we use the Adam optimizer and the batch size equals to $1$. Since the target data size is smaller than the source data, each target data is loaded randomly multiple times in each epoch during training.
For the weighting of loss functions, we follow the common strategy as \cite{ganin2015unsupervised,ganin2016domain} to gradually increase $\beta_l$ and $\beta_g$ from $0$ to $1$. the weighting $\alpha$ for smoothness loss is $0.15$ as in \cite{farha2019ms} and $\mu$ is chosen as $1 \times 10^{-4}$ via the grid-search.

\subsection{Experimental Results}
We first compare with the baseline model MS-TCN~\cite{farha2019ms} to see how our approaches effectively utilize the unlabeled videos for action segmentation. ``Source only" means the model is trained only with source labeled videos. And then we compare the proposed approach to
the state-of-the-art methods on all three datasets.

\noindent\textbf{Local Temporal Domain Adaptation.}
By integrating domain classifiers with frame-level features $\textit{\textbf{f}}$, the results on all three datasets with respect to all the metrics are improved significantly, as shown in the row ``DA (\textit{L})" in \Cref{table:exp_local_global_DA}. For example, on the GTEA dataset, our approach outperforms the baseline by 4.6\% for F1$@$50, 5.5\% for the edit score and 3.8\% for the frame-wise accuracy. Although ``DA (\textit{L})" mainly works on the frame-level features, they are learned using the context from neighbor frames, so they still contain temporal information, which is critical to diminish the temporal variations for actions across domains.

\begin{table}[!t]
\centering
\small
    \begin{tabular}{c|ccc|c|c} 
    \hline
    \textbf{GTEA} & \multicolumn{3}{c|}{F1$@\{10,25,50\}$} & Edit & Acc \\ \hline
    Source only (MS-TCN) & 85.8 & 83.4 & 69.8 & 79.0 & 76.3 \\ 
    DA (\textit{L}) & 89.6 & 87.9 & 74.4 & 84.5 & 80.1 \\ 
    DA (\textit{L + G}) & 90.0 & 88.6 & 74.9 & 85.6 & 79.6 \\ 
    DA (\textit{L + G + A}) & 90.5 & 88.4 & 76.2 & 85.8 & 80.0 \\ \hline
    \hline
    \textbf{50Salads} & \multicolumn{3}{c|}{F1$@\{10,25,50\}$} & Edit & Acc \\ \hline
    Source only (MS-TCN) & 76.3 & 74.0 & 64.5 & 67.9 & 80.7 \\ 
    DA (\textit{L}) & 79.2 & 77.8 & 70.3 & 72.0 & 82.8 \\ 
    DA (\textit{L + G}) & 80.2 & 78.4 & 70.6 & 73.4 & 82.2 \\ 
    DA (\textit{L + G + A}) & 82.0 & 80.1 & 72.5 & 75.2 & 83.2 \\ \hline
    \hline
    \textbf{Breakfast} & \multicolumn{3}{c|}{F1$@\{10,25,50\}$} & Edit & Acc \\ \hline
    Source only (MS-TCN) & 52.6 & 48.1 & 37.9 & 61.7 & 66.3 \\ 
    DA (\textit{L}) & 72.8 & 67.8 & 55.1 & 71.7 & 70.3 \\ 
    DA (\textit{L + G}) & 72.6 & 66.9 & 54.3 & 72.6 & 69.2 \\ 
    DA (\textit{L + G + A}) & 74.2 & 68.6 & 56.5 & 73.6 & 71.0 \\ \hline
    \end{tabular}
\caption{The experimental results for all our approaches on GTEA, 50Salads, and the Breakfast dataset (\textit{L}: local temporal DA, \textit{G}: global temporal DA without domain attention, \textit{A}: domain attention mechanism).} 
\label{table:exp_local_global_DA}
\end{table}

\noindent\textbf{Mixed Temporal Domain Adaptation.}
Despite the improvement from local temporal DA, the temporal receptive fields of frame-level features are still not guaranteed to cover the whole video length. Therefore, we aggregate frame-level features to generate a video-level feature for each video and apply additional domain classifier on it. However, aggregating frames by temporal pooling without considering the importance of each frame does not ensure better performance, especially for the Breakfast dataset, which contains much higher domain discrepancy than the other two. The F1 score and frame-wise accuracy both have slightly worse results, as shown in the row ``DA (\textit{L + G})" in \Cref{table:exp_local_global_DA}. Therefore, we apply the domain attention mechanism to aggregate frames more effectively, leading to better global temporal DA performance. For example, on the Breakfast dataset, ``DA (\textit{L + G + A})" outperforms ``DA (\textit{L})" by 1.4\% for F1$@$50, 1.9\% for the edit score and 0.7\% for the frame-wise accuracy, as shown in \Cref{table:exp_local_global_DA}.

Our final method ``DA (\textit{L + G + A})", which is also MTDA, outperforms the baseline by large margins (e.g. 6.4\% for F1$@$50, 6.8\% for the edit score and 3.7\% for the frame-wise accuracy on GTEA; 8.0\% for F1$@$50, 7.3\% for the edit score and 2.5\% for the frame-wise accuracy on 50Salads), as demonstrated in \Cref{table:exp_local_global_DA}.

\noindent\textbf{Comparison with the State-of-the-Art.}
Here we compare the proposed MTDA to the state-of-the-art methods, and MTDA outperforms all the previous methods on the three datasets with respect to three evaluation metrics: F1 score, edit distance, and frame-wise accuracy, as shown in \Cref{table:SOTA}.

For the GTEA dataset, the authors of MS-TCN~\cite{farha2019ms} also fine-tune the I3D features to improve the performance (e.g. from 85.8\% to 87.5\% for F1$@10$). Our MTDA outperforms the fine-tuned MS-TCN even without any fine-tuning process since we learn the temporal features more effectively from unlabeled videos, which is more important for action segmentation.

cGAN~\cite{gammulle2019coupled} utilizes supplementary modalities including depth maps and optical flow with an auxiliary network. cGAN outperforms MS-TCN in terms of the F1 score and edit score on the 50Salads dataset. Our MTDA outperforms cGAN, indicating more effective way to utilizes auxiliary data. 

For the Breakfast dataset, the authors of MS-TCN~\cite{farha2019ms} also use the improved dense trajectories (IDT) features, which encode only motion information and outperform the I3D features since the encoded spatial information is not the critical factor for the Breakfast dataset. Our MTDA outperforms the IDT-version of MS-TCN by a large margin with the same I3D features. This shows that our DATP module effectively aggregate frames by considering the temporal structure for action segmentation.

\begin{table}[!t]
\centering
\small
    \begin{tabular}{c|ccc|c|c} 
    \hline
    \textbf{GTEA} & \multicolumn{3}{c|}{F1$@\{10,25,50\}$} & Edit & Acc \\ \hline
    ST-CNN~\cite{lea2016segmental} & 58.7 & 54.4 & 41.9 & 49.1 & 60.6 \\ 
    Bi-LSTM~\cite{singh2016multi} & 66.5 & 59.0 & 43.6 & - & 55.5 \\ 
    ED-TCN~\cite{lea2017temporal} & 72.2 & 69.3 & 56.0 & - & 64.0 \\ 
    TricorNet~\cite{ding2017tricornet} & 76.0 & 71.1 & 59.2 & - & 64.8 \\ 
    TDRN~\cite{lei2018temporal} & 79.2 & 74.4 & 62.7 & 74.1 & 70.1 \\ 
    cGAN~\cite{gammulle2019coupled} & 80.1 & 77.9 & 69.1 & 78.1 & 78.5 \\ 
    MS-TCN~\cite{farha2019ms} & 85.8 & 83.4 & 69.8 & 79.0 & 76.3 \\ 
    MS-TCN (FT)~\cite{farha2019ms} & 87.5 & 85.4 & 74.6 & 81.4 & 79.2 \\ \hline
    \textbf{MTDA} & \textbf{90.5} & \textbf{88.4} & \textbf{76.2} & \textbf{85.8} & \textbf{80.0} \\ \hline
    \hline
    \textbf{50Salads} & \multicolumn{3}{c|}{F1$@\{10,25,50\}$} & Edit & Acc \\ \hline
    IDT+LM~\cite{richard2016temporal} & 44.4 & 38.9 & 27.8 & 45.8 & 48.7 \\ 
    Bi-LSTM~\cite{singh2016multi} & 62.6 & 58.3 & 47.0 & 55.6 & 55.7 \\ 
    ST-CNN~\cite{lea2016segmental} & 55.9 & 49.6 & 37.1 & 45.9 & 59.4 \\ 
    ED-TCN~\cite{lea2017temporal} & 68.0 & 63.9 & 52.6 & 59.8 & 64.7 \\ 
    TricorNet~\cite{ding2017tricornet} & 70.1 & 67.2 & 56.6 & 62.8 & 67.5 \\ 
    TDRN~\cite{lei2018temporal} & 72.9 & 68.5 & 57.2 & 66.0 & 68.1 \\ 
    MS-TCN~\cite{farha2019ms} & 76.3 & 74.0 & 64.5 & 67.9 & 80.7 \\ 
    cGAN~\cite{gammulle2019coupled} & 80.1 & 78.7 & 71.1 & 76.9 & 74.5 \\ \hline 
    \textbf{MTDA} & \textbf{82.0} & \textbf{80.1} & \textbf{72.5} & \textbf{75.2} & \textbf{83.2} \\ \hline
    \hline
    \textbf{Breakfast} & \multicolumn{3}{c|}{F1$@\{10,25,50\}$} & Edit & Acc \\ \hline
    ED-TCN~\cite{lea2017temporal} & - & - & - & - & 43.3 \\ 
    HTK~\cite{kuehne2017weakly} & - & - & - & - & 50.7 \\ 
    TCFPN~\cite{ding2018weakly} & - & - & - & - & 52.0 \\ 
    HTK (64)~\cite{kuehne2016end} & - & - & - & - & 56.3 \\ 
    GRU~\cite{richard2017weakly} & - & - & - & - & 60.6 \\ 
    MS-TCN~\cite{farha2019ms} & 52.6 & 48.1 & 37.9 & 61.7 & 66.3 \\ 
    MS-TCN (IDT)~\cite{farha2019ms} & 58.2 & 52.9 & 40.8 & 61.4 & 65.1 \\ \hline
    \textbf{MTDA} & \textbf{74.2} & \textbf{68.6} & \textbf{56.5} & \textbf{73.6} & \textbf{71.0} \\ \hline
    \end{tabular}
\caption{Comparison with the state-of-the-art on GTEA, 50Salads,
and the Breakfast dataset.} 
\label{table:SOTA}
\end{table}

\begin{table}[!t]
\centering
\small
    \begin{tabular}{c|ccc|c|c} 
    \hline
     & \multicolumn{3}{c|}{F1$@\{10,25,50\}$} & Edit & Acc \\ \hline
    Source only & 85.8 & 83.4 & 69.8 & 79.0 & 76.3 \\ \hline
    $\{S1\}$ & 88.6 & 86.2 & 73.6 & 84.2 & 78.7 \\  
    $\{S2\}$ & 89.1 & 87.2 & \textbf{74.4} & 84.3 & 79.1 \\  
    $\{S3\}$ & 89.2 & 87.3 & 72.3 & 83.8 & 78.9 \\  
    $\{S4\}$ & 88.1 & 86.4 & 73.0 & 83.0 & 78.8 \\  \hline
    $\{S1, S2\}$ & 89.0 & 85.8 & 73.5 & \textbf{84.8} & 79.5 \\  
    $\{S2, S3\}$ & \textbf{89.6} & \textbf{87.9} & \textbf{74.4} & 84.5 & \textbf{80.1} \\  
    $\{S3, S4\}$ & 88.3 & 86.8 & 73.9 & 83.6 & 78.6 \\  \hline
    \end{tabular}
\caption{The experimental results for the integration of $G_{ld}$ and different stages of MS-TCN~\cite{farha2019ms} on the GTEA dataset. $\{S_n\}$ means adding $G_{ld}$ to the $n$th stage. The stages with smaller $n$ are closer to inputs.} 
\label{table:stage_analysis}
\end{table}

\begin{figure*}[!t]
\centering
\includegraphics[width=\textwidth]{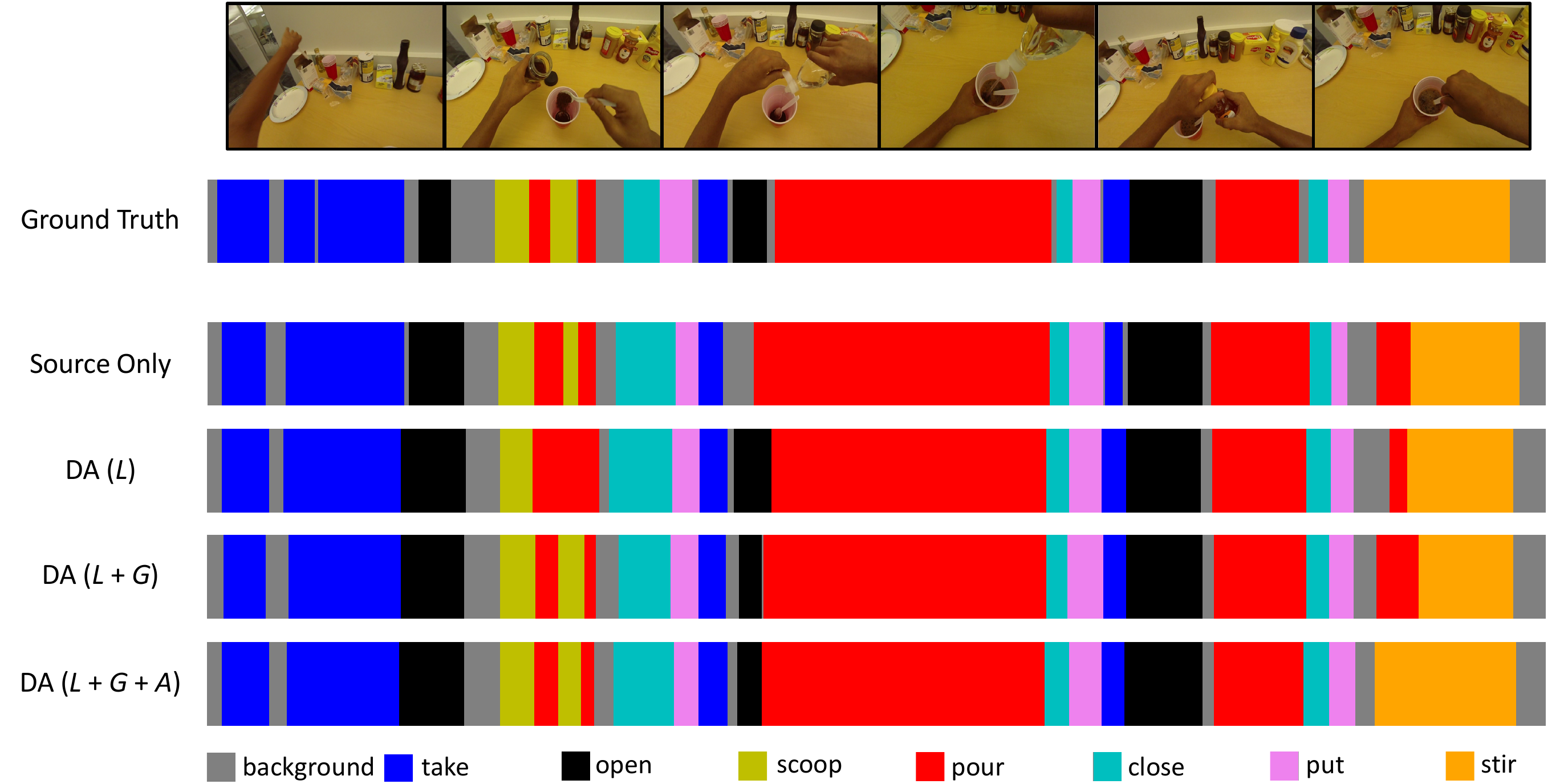}
\caption{The qualitative results of temporal action segmentation on GTEA for the activity \textit{CofHoney}. The video snapshots are shown in the first row in a temporal order (from left to right). ``Source only" refers to the baseline model MS-TCN~\cite{farha2019ms}.}
\label{fig:qualitative_results}
\end{figure*}

\subsection{Ablation Study and Analysis}
\noindent\textbf{Integration of $G_{ld}$ and Stages.} \label{sec:stage_analysis}
Since we use Multi-stage TCN~\cite{farha2019ms} as our baseline model and develop our approaches upon it, it raises a question: \textit{How to effectively perform DA by integrating the domain classifiers to a multi-stage architecture?}
Our architecture contains four stages as in \cite{farha2019ms}. First, we integrate $G_{ld}$ into one stage and the results are demonstrated in \Cref{table:stage_analysis}. The results show that the best performance happens when $G_{ld}$s are integrated into middle stages, such as $S2$ or $S3$. $S1$ is not a good choice for DA because of two reasons: 1) $S1$ corresponds to low-level and transferable features with less discriminability where DA shows limited effects~\cite{long2015learning}. 2) $S1$ capture less temporal information from neighbor frames, representing less temporal receptive fields, which is critical for action segmentation. However, higher stages (e.g. $S4$) are not always better. We conjecture that it is because higher stages are used to refine the prediction. They may affect the semantic structure of feature representations, which is important to DA. In our case, integrating $G_{ld}$ into $S2$ provides the best overall performance.

We also add multiple domain classifiers to adjacent stages. However, multi-stage DA does not always guarantee improved performance. For example, $\{S1, S2\}$ has worse results than $\{S2\}$ in terms of F1$@\{10,25,50\}$. Since $\{S2\}$ and $\{S3\}$ provide the best single-stage DA performance, we use $\{S2, S3\}$, which performs the best, as the final model for all our approaches in all the experiments.

\noindent\textbf{Qualitative results.}
In addition to evaluating the quantitative performance using the above metrics, it is also common to evaluate the qualitative performance to ensure that the prediction results are aligned with human vision.
Here we compare our approaches with the baseline model MS-TCN~\cite{farha2019ms} and the ground truth, as shown in \Cref{fig:qualitative_results}.
MS-TCN fails to predict \textit{open} before the long \textit{pour} action in the middle part of the video, and falsely predict \textit{pour} before \textit{stir} in the end of the video, as shown in the ``Source only" row. With local and global temporal DA, our approach can detect all the actions happened in the video, as shown in the row ``DA (\textit{L + G})". Finally, with the domain attention mechanism, our proposed MTDA also removes the falsely predicted action \textit{pour}. For more qualitative results, please refer to the supplementary material.

\section{Conclusion and Future Work}
In this paper, we consider action segmentation as a DA problem and reduce the domain discrepancy by performing unsupervised DA with auxiliary unlabeled videos. 
To diminish domain discrepancy for both the spatial and temporal directions, we propose \textbf{Mixed Temporal Domain Adaptation (MTDA)} to jointly align frame- and video-level embedded feature spaces across domains, and further integrate with the domain attention mechanism to focus on aligning the frame-level features with higher domain discrepancy, leading to more effective domain adaptation. 
The comprehensive experiment results validate that our approach outperforms all the previous state-of-the-art methods. Our approach can adapt models effectively by using auxiliary unlabeled videos, leading to further possible applications to large-scale problems, such as video surveillance and human activity analysis.
For the future work, we would like to develop DA approaches with self-supervised learning to reduce the need of additional unlabeled videos.


\section{Appendix}
\subsection{More Qualitative results}

\begin{figure*}[!t]
  \begin{subfigure}[b]{\textwidth}
    \includegraphics[width=\textwidth]{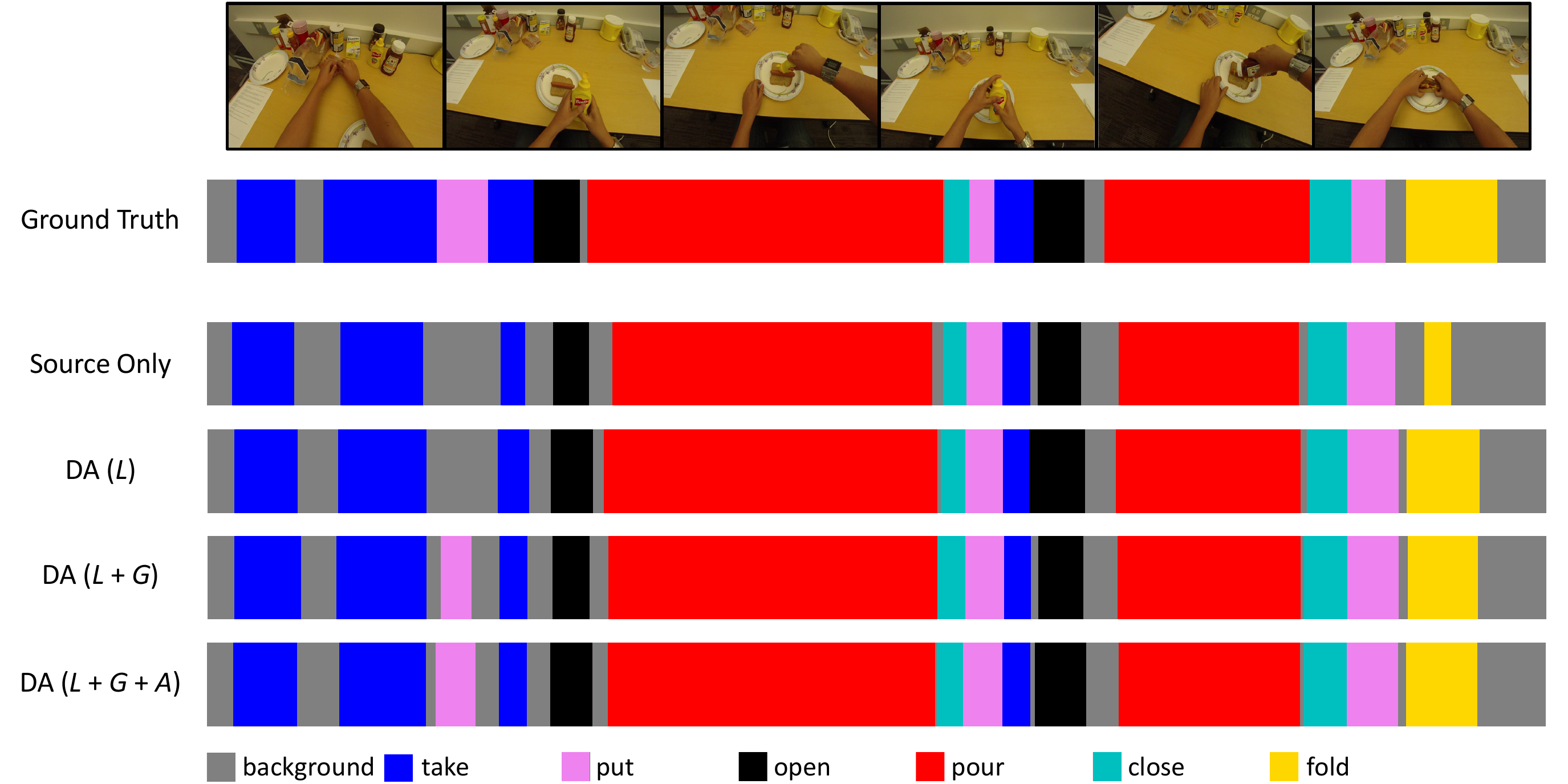}
    \caption{\textit{Hotdog}}
    \label{fig:Hotdog_supp}
  \end{subfigure}
  \begin{subfigure}[b]{\textwidth}
    \includegraphics[width=\textwidth]{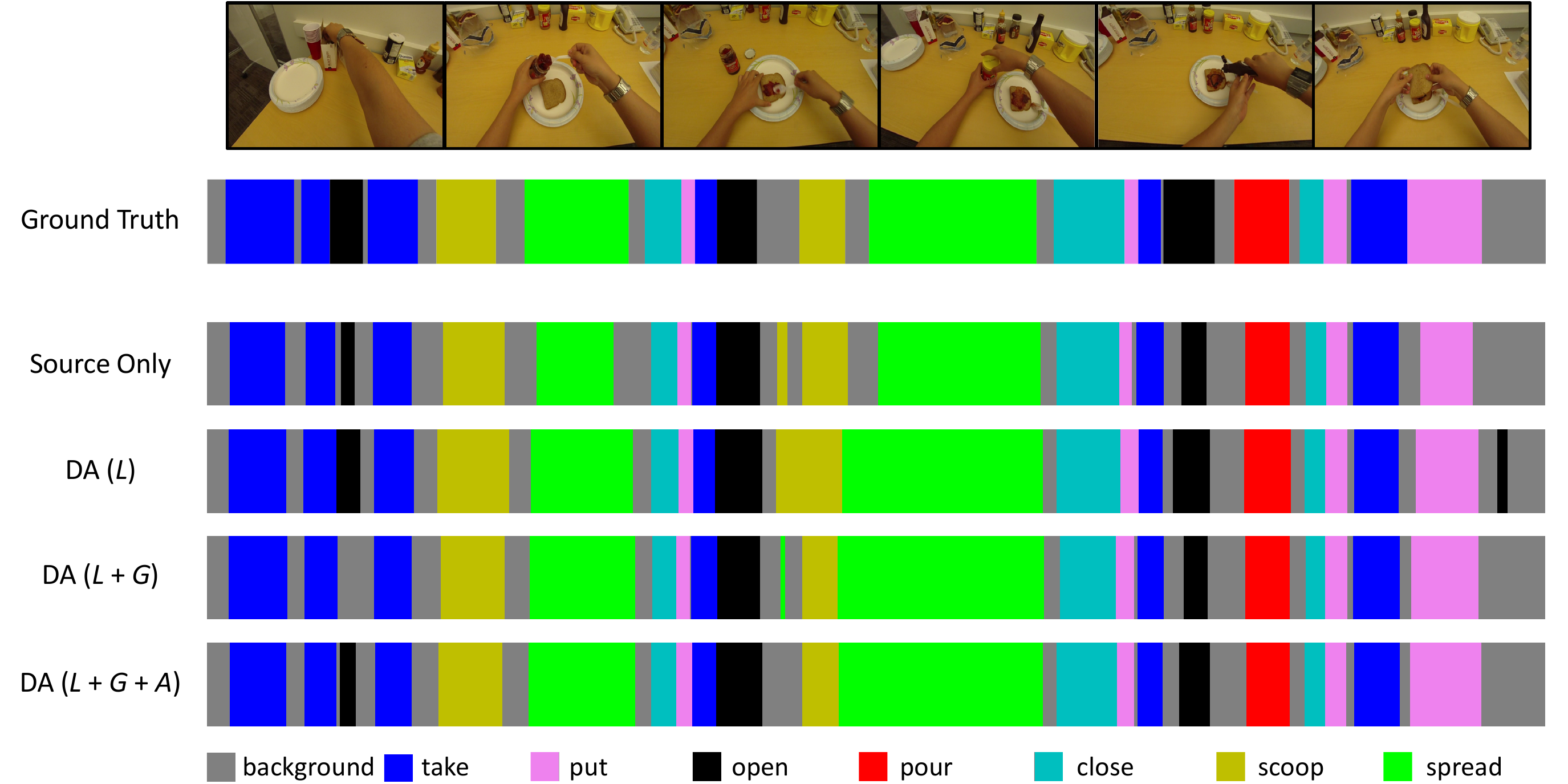}
    \caption{\textit{Pealate}}
    \label{fig:Pealate_supp}
  \end{subfigure}
\caption{The qualitative results of temporal action segmentation on GTEA for the activity (a) \textit{Hotdog} and (b) \textit{Pealate}. The video snapshots are shown in the first row in a temporal order (from left to right). ``Source only" refers to the baseline model MS-TCN~\cite{farha2019ms}.}
\label{fig:qualitative_results_gtea_supp}
\end{figure*}

\begin{figure*}[!t]
  \begin{subfigure}[b]{\textwidth}
    \includegraphics[width=\textwidth]{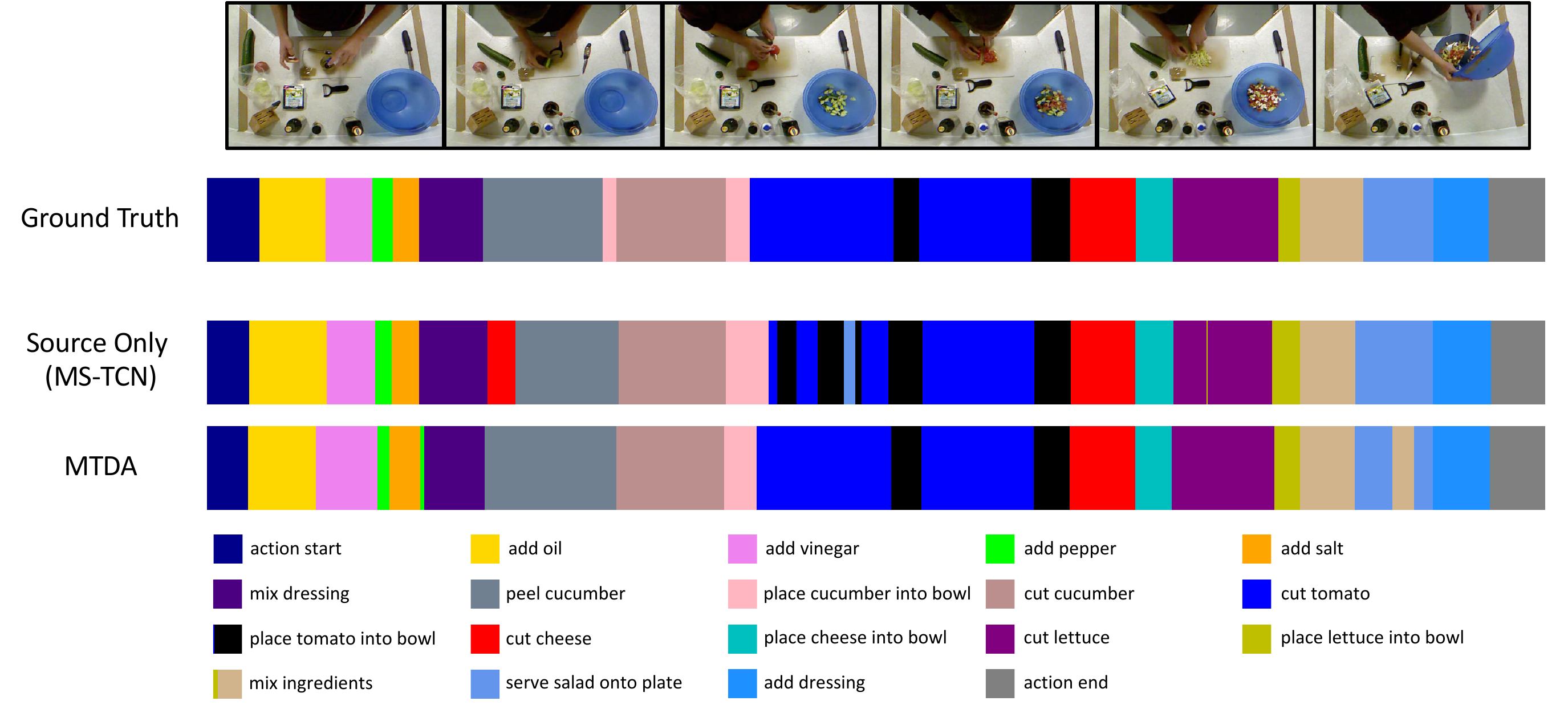}
    \caption{\textit{50Salads}}
    \label{fig:50Salads_supp}
  \end{subfigure}
  \begin{subfigure}[b]{\textwidth}
    \includegraphics[width=\textwidth]{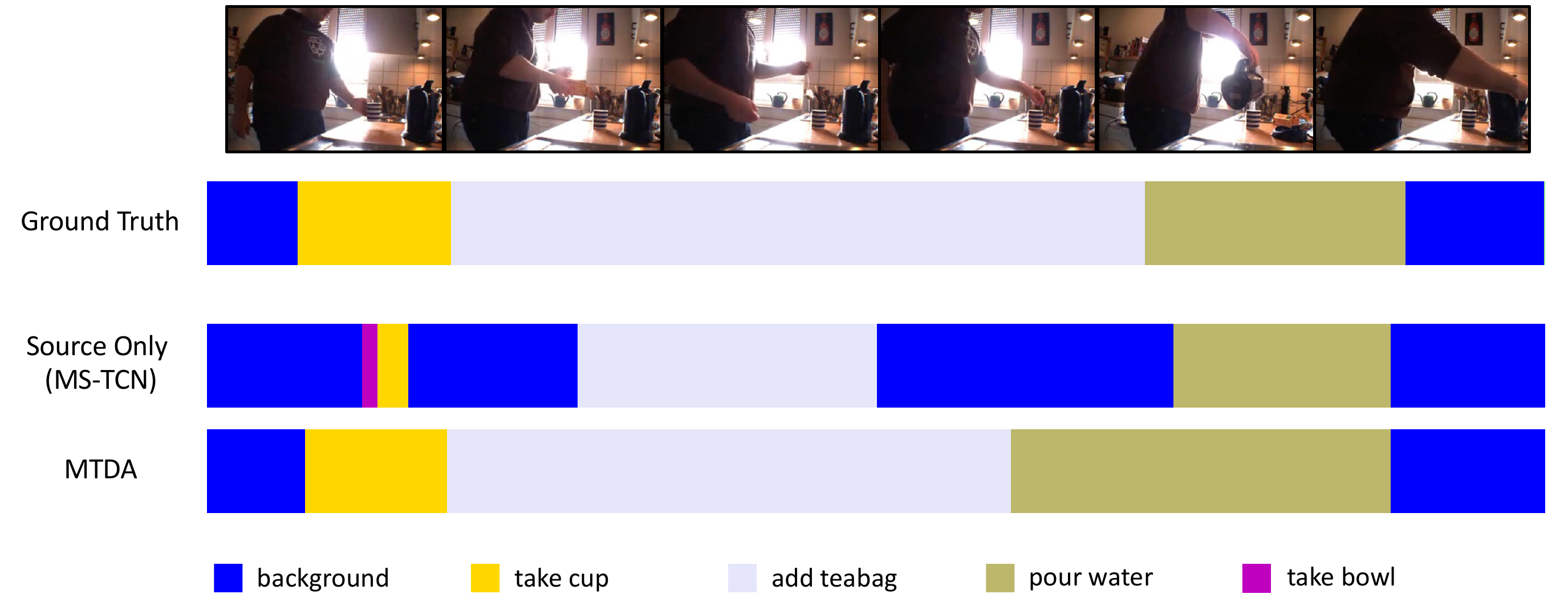}
    \caption{\textit{Breakfast}}
    \label{fig:Breakfast_supp}
  \end{subfigure}
\caption{The qualitative results of temporal action segmentation on (a) \textit{50salads} and (b) \textit{Breakfast}. The video snapshots are shown in the first row in a temporal order (from left to right). ``Source only" refers to the baseline model MS-TCN~\cite{farha2019ms}.}
\label{fig:qualitative_results_50salads_breakfast_supp}
\end{figure*}

Here we show more examples to compare our approaches with the baseline model MS-TCN~\cite{farha2019ms} and the ground truth, as shown in \Cref{fig:qualitative_results_gtea_supp,fig:qualitative_results_50salads_breakfast_supp}.

\noindent\textbf{GTEA dataset.} 
For the example \textit{Hotdog} (\Cref{fig:Hotdog_supp}), MS-TCN fails to predict the \textit{put} action before the \textit{take} action in the early part of the video, and the predicted \textit{fold} action in the end of the video has much shorter time duration than it should be, as shown in the ``Source only" row. With local and global temporal DA, our approach can detect all the actions with proper time duration, as shown in the row ``DA (\textit{L + G})", and the domain attention mechanism further helps refine the time duration for each predicted action. For another example \textit{Pealate} (\Cref{fig:Pealate_supp}), our final approach ``DA (\textit{L + G + A})" also produces the best action segmentation result, which is the closest to the ground truth, compared with the baseline and other methods.

\noindent\textbf{50Salads and Breakfast datasets.}
In addition to the GTEA dataset, we also evaluate the qualitative performance on another two challenger datasets: \textit{50Salads} and \textit{Breakfast}, as demonstrated in \Cref{fig:qualitative_results_50salads_breakfast_supp}.
The 50Salads dataset is challenging since each video is long and contains around 20 fine-grained action classes. While MS-TCN confuses with some similar classes like \textit{``cut tomato"} and \textit{``place tomato into bowl"}, our approach can produce smooth temporal segmentation, as shown in \Cref{fig:50Salads_supp}.
The Breakfast dataset is challenging because of high spatio-temporal variations among videos since the videos are recorded in 18 different kitchens with 52 subjects. The total number of action classes is also much larger than the other two datasets. \Cref{fig:Breakfast_supp} shows that MS-TCN falsely classifies \textit{``take cup"} as \textit{``take bowl"} and unable to detect the \textit{``add teabag"} action for a long time. However, our proposed MTDA can continuously detect actions in the video without gaps along the temporal direction.


{\small
\bibliographystyle{ieee}
\bibliography{egbib}

\begin{thebibliography}{10}\itemsep=-1pt

\bibitem{carreira2017quo}
J.~Carreira and A.~Zisserman.
\newblock Quo vadis, action recognition? a new model and the kinetics dataset.
\newblock In {\em IEEE conference on Computer Vision and Pattern Recognition
  (CVPR)}, 2017.

\bibitem{chen2019temporal}
M.-H. Chen, Z.~Kira, G.~AlRegib, J.~Woo, R.~Chen, and J.~Zheng.
\newblock Temporal attentive alignment for large-scale video domain adaptation.
\newblock In {\em IEEE International Conference on Computer Vision (ICCV)},
  2019.

\bibitem{csurka2017comprehensive}
G.~Csurka.
\newblock A comprehensive survey on domain adaptation for visual applications.
\newblock In {\em Domain Adaptation in Computer Vision Applications}, pages
  1--35. Springer, 2017.

\bibitem{ding2017tricornet}
L.~Ding and C.~Xu.
\newblock Tricornet: A hybrid temporal convolutional and recurrent network for
  video action segmentation.
\newblock {\em arXiv preprint arXiv:1705.07818}, 2017.

\bibitem{ding2018weakly}
L.~Ding and C.~Xu.
\newblock Weakly-supervised action segmentation with iterative soft boundary
  assignment.
\newblock In {\em IEEE Conference on Computer Vision and Pattern Recognition
  (CVPR)}, 2018.

\bibitem{donahue2014decaf}
J.~Donahue, Y.~Jia, O.~Vinyals, J.~Hoffman, N.~Zhang, E.~Tzeng, and T.~Darrell.
\newblock Decaf: A deep convolutional activation feature for generic visual
  recognition.
\newblock In {\em International Conference on Machine Learning (ICML)}, 2014.

\bibitem{farha2019ms}
Y.~A. Farha and J.~Gall.
\newblock Ms-tcn: Multi-stage temporal convolutional network for action
  segmentation.
\newblock In {\em IEEE Conference on Computer Vision and Pattern Recognition
  (CVPR)}, 2019.

\bibitem{fathi2011learning}
A.~Fathi, X.~Ren, and J.~M. Rehg.
\newblock Learning to recognize objects in egocentric activities.
\newblock In {\em IEEE Conference on Computer Vision and Pattern Recognition
  (CVPR)}, 2011.

\bibitem{gammulle2019coupled}
H.~Gammulle, T.~Fernando, S.~Denman, S.~Sridharan, and C.~Fookes.
\newblock Coupled generative adversarial network for continuous fine-grained
  action segmentation.
\newblock In {\em IEEE Winter Conference on Applications of Computer Vision
  (WACV)}, 2019.

\bibitem{ganin2015unsupervised}
Y.~Ganin and V.~Lempitsky.
\newblock Unsupervised domain adaptation by backpropagation.
\newblock In {\em International Conference on Machine Learning (ICML)}, 2015.

\bibitem{ganin2016domain}
Y.~Ganin, E.~Ustinova, H.~Ajakan, P.~Germain, H.~Larochelle, F.~Laviolette,
  M.~Marchand, and V.~Lempitsky.
\newblock Domain-adversarial training of neural networks.
\newblock {\em The Journal of Machine Learning Research (JMLR)},
  17(1):2096--2030, 2016.

\bibitem{goodfellow2014generative}
I.~Goodfellow, J.~Pouget-Abadie, M.~Mirza, B.~Xu, D.~Warde-Farley, S.~Ozair,
  A.~Courville, and Y.~Bengio.
\newblock Generative adversarial nets.
\newblock In {\em Advances in Neural Information Processing Systems (NIPS)},
  2014.

\bibitem{jamal2018deep}
A.~Jamal, V.~P. Namboodiri, D.~Deodhare, and K.~Venkatesh.
\newblock Deep domain adaptation in action space.
\newblock In {\em British Machine Vision Conference (BMVC)}, 2018.

\bibitem{kuehne2014language}
H.~Kuehne, A.~Arslan, and T.~Serre.
\newblock The language of actions: Recovering the syntax and semantics of
  goal-directed human activities.
\newblock In {\em IEEE Conference on Computer Vision and Pattern Recognition
  (CVPR)}, 2014.

\bibitem{kuehne2016end}
H.~Kuehne, J.~Gall, and T.~Serre.
\newblock An end-to-end generative framework for video segmentation and
  recognition.
\newblock In {\em IEEE Winter Conference on Applications of Computer Vision
  (WACV)}, 2016.

\bibitem{kuehne2017weakly}
H.~Kuehne, A.~Richard, and J.~Gall.
\newblock Weakly supervised learning of actions from transcripts.
\newblock {\em Computer Vision and Image Understanding (CVIU)}, 163:78--89,
  2017.

\bibitem{lea2017temporal}
C.~Lea, M.~D. Flynn, R.~Vidal, A.~Reiter, and G.~D. Hager.
\newblock Temporal convolutional networks for action segmentation and
  detection.
\newblock In {\em IEEE Conference on Computer Vision and Pattern Recognition
  (CVPR)}, 2017.

\bibitem{lea2016segmental}
C.~Lea, A.~Reiter, R.~Vidal, and G.~D. Hager.
\newblock Segmental spatiotemporal cnns for fine-grained action segmentation.
\newblock In {\em European Conference on Computer Vision (ECCV)}, 2016.

\bibitem{lei2018temporal}
P.~Lei and S.~Todorovic.
\newblock Temporal deformable residual networks for action segmentation in
  videos.
\newblock In {\em IEEE Conference on Computer Vision and Pattern Recognition
  (CVPR)}, 2018.

\bibitem{li2018adaptive}
Y.~Li, N.~Wang, J.~Shi, X.~Hou, and J.~Liu.
\newblock Adaptive batch normalization for practical domain adaptation.
\newblock {\em Pattern Recognition}, 80:109--117, 2018.

\bibitem{li2017revisiting}
Y.~Li, N.~Wang, J.~Shi, J.~Liu, and X.~Hou.
\newblock Revisiting batch normalization for practical domain adaptation.
\newblock In {\em International Conference on Learning Representations Workshop
  (ICLRW)}, 2017.

\bibitem{long2015learning}
M.~Long, Y.~Cao, J.~Wang, and M.~Jordan.
\newblock Learning transferable features with deep adaptation networks.
\newblock In {\em International Conference on Machine Learning (ICML)}, 2015.

\bibitem{long2016unsupervised}
M.~Long, H.~Zhu, J.~Wang, and M.~I. Jordan.
\newblock Unsupervised domain adaptation with residual transfer networks.
\newblock In {\em Advances in Neural Information Processing Systems (NIPS)},
  2016.

\bibitem{long2017deep}
M.~Long, H.~Zhu, J.~Wang, and M.~I. Jordan.
\newblock Deep transfer learning with joint adaptation networks.
\newblock In {\em International Conference on Machine Learning (ICML)}, 2017.

\bibitem{ma2019ts}
C.-Y. Ma, M.-H. Chen, Z.~Kira, and G.~AlRegib.
\newblock Ts-lstm and temporal-inception: Exploiting spatiotemporal dynamics
  for activity recognition.
\newblock {\em Signal Processing: Image Communication}, 71:76--87, 2019.

\bibitem{ma2018attend}
C.-Y. Ma, A.~Kadav, I.~Melvin, Z.~Kira, G.~AlRegib, and H.~P. Graf.
\newblock Attend and interact: Higher-order object interactions for video
  understanding.
\newblock In {\em IEEE conference on Computer Vision and Pattern Recognition
  (CVPR)}, 2018.

\bibitem{newell2016stacked}
A.~Newell, K.~Yang, and J.~Deng.
\newblock Stacked hourglass networks for human pose estimation.
\newblock In {\em EEuropean Conference on Computer Vision (ECCV)}, 2016.

\bibitem{oord2016wavenet}
A.~v.~d. Oord, S.~Dieleman, H.~Zen, K.~Simonyan, O.~Vinyals, A.~Graves,
  N.~Kalchbrenner, A.~Senior, and K.~Kavukcuoglu.
\newblock Wavenet: A generative model for raw audio.
\newblock {\em arXiv preprint arXiv:1609.03499}, 2016.

\bibitem{pan2010survey}
S.~J. Pan, Q.~Yang, et~al.
\newblock A survey on transfer learning.
\newblock {\em IEEE Transactions on Knowledge and Data Engineering (TKDE)},
  22(10):1345--1359, 2010.

\bibitem{paszke2017automatic}
A.~Paszke, S.~Gross, S.~Chintala, G.~Chanan, E.~Yang, Z.~DeVito, Z.~Lin,
  A.~Desmaison, L.~Antiga, and A.~Lerer.
\newblock Automatic differentiation in pytorch.
\newblock In {\em Advances in Neural Information Processing Systems Workshop
  (NIPSW)}, 2017.

\bibitem{richard2016temporal}
A.~Richard and J.~Gall.
\newblock Temporal action detection using a statistical language model.
\newblock In {\em IEEE Conference on Computer Vision and Pattern Recognition
  (CVPR)}, 2016.

\bibitem{richard2017weakly}
A.~Richard, H.~Kuehne, and J.~Gall.
\newblock Weakly supervised action learning with rnn based fine-to-coarse
  modeling.
\newblock In {\em IEEE Conference on Computer Vision and Pattern Recognition
  (CVPR)}, 2017.

\bibitem{saito2018maximum}
K.~Saito, K.~Watanabe, Y.~Ushiku, and T.~Harada.
\newblock Maximum classifier discrepancy for unsupervised domain adaptation.
\newblock In {\em IEEE conference on Computer Vision and Pattern Recognition
  (CVPR)}, 2018.

\bibitem{singh2016multi}
B.~Singh, T.~K. Marks, M.~Jones, O.~Tuzel, and M.~Shao.
\newblock A multi-stream bi-directional recurrent neural network for
  fine-grained action detection.
\newblock In {\em IEEE Conference on Computer Vision and Pattern Recognition
  (CVPR)}, 2016.

\bibitem{stein2013combining}
S.~Stein and S.~J. McKenna.
\newblock Combining embedded accelerometers with computer vision for
  recognizing food preparation activities.
\newblock In {\em ACM international joint conference on Pervasive and
  ubiquitous computing (UbiComp)}, 2013.

\bibitem{sultani2014human}
W.~Sultani and I.~Saleemi.
\newblock Human action recognition across datasets by foreground-weighted
  histogram decomposition.
\newblock In {\em IEEE conference on Computer Vision and Pattern Recognition
  (CVPR)}, 2014.

\bibitem{sun2016deep}
B.~Sun and K.~Saenko.
\newblock Deep coral: Correlation alignment for deep domain adaptation.
\newblock In {\em European Conference on Computer Vision Workshop (ECCVW)},
  2016.

\bibitem{tran2015learning}
D.~Tran, L.~Bourdev, R.~Fergus, L.~Torresani, and M.~Paluri.
\newblock Learning spatiotemporal features with 3d convolutional networks.
\newblock In {\em IEEE International Conference on Computer Vision (ICCV)},
  2015.

\bibitem{tzeng2017adversarial}
E.~Tzeng, J.~Hoffman, K.~Saenko, and T.~Darrell.
\newblock Adversarial discriminative domain adaptation.
\newblock In {\em IEEE Conference on Computer Vision and Pattern Recognition
  (CVPR)}, 2017.

\bibitem{wang2018non}
X.~Wang, R.~Girshick, A.~Gupta, and K.~He.
\newblock Non-local neural networks.
\newblock In {\em IEEE conference on Computer Vision and Pattern Recognition
  (CVPR)}, 2018.

\bibitem{wang2019transferable}
X.~Wang, L.~Li, W.~Ye, M.~Long, and J.~Wang.
\newblock Transferable attention for domain adaptation.
\newblock In {\em AAAI Conference on Artificial Intelligence (AAAI)}, 2019.

\bibitem{wei2016convolutional}
S.-E. Wei, V.~Ramakrishna, T.~Kanade, and Y.~Sheikh.
\newblock Convolutional pose machines.
\newblock In {\em IEEE Conference on Computer Vision and Pattern Recognition
  (CVPR)}, 2016.

\bibitem{xu2016dual}
T.~Xu, F.~Zhu, E.~K. Wong, and Y.~Fang.
\newblock Dual many-to-one-encoder-based transfer learning for cross-dataset
  human action recognition.
\newblock {\em Image and Vision Computing}, 55:127--137, 2016.

\bibitem{yan2017mind}
H.~Yan, Y.~Ding, P.~Li, Q.~Wang, Y.~Xu, and W.~Zuo.
\newblock Mind the class weight bias: Weighted maximum mean discrepancy for
  unsupervised domain adaptation.
\newblock In {\em IEEE Conference on Computer Vision and Pattern Recognition
  (CVPR)}, 2017.

\bibitem{zellinger2017central}
W.~Zellinger, T.~Grubinger, E.~Lughofer, T.~Natschl{\"a}ger, and
  S.~Saminger-Platz.
\newblock Central moment discrepancy (cmd) for domain-invariant representation
  learning.
\newblock In {\em International Conference on Learning Representations (ICLR)},
  2017.

\bibitem{zhang2018collaborative}
W.~Zhang, W.~Ouyang, W.~Li, and D.~Xu.
\newblock Collaborative and adversarial network for unsupervised domain
  adaptation.
\newblock In {\em IEEE Conference on Computer Vision and Pattern Recognition
  (CVPR)}, 2018.

\end{thebibliography}
}

\end{document}